%% file: main.tex
\documentclass[letterpaper, conference]{IEEEtran}
\pdfoutput=1
\IEEEoverridecommandlockouts

\usepackage[cmex10]{amsmath}
\usepackage{cite}
\usepackage{xcolor}

\usepackage{amssymb}
\usepackage{amsfonts}
\usepackage{pifont}

\usepackage{enumitem}
\usepackage{path}
\usepackage{verbatim}
\usepackage{algorithm}
\usepackage{algorithmic}
\usepackage{framed}
\usepackage{footnote}
\usepackage{color}
\usepackage{amsmath}
\usepackage{cases}
\usepackage{caption}
\usepackage{subcaption}
\usepackage{theorem}
\usepackage{xurl}
\usepackage{acro}
\usepackage{eurosym}
\usepackage{multirow}
\usepackage{makecell}

\usepackage[normalem]{ulem}
\usepackage{tikz}

\usepackage{tabularx}
\usepackage{booktabs}
\usepackage[colorlinks=true, allcolors=black]{hyperref}
\usepackage{threeparttable}

\usepackage{graphicx}

\usepackage{pgfplots}
\usepackage{pgfplotstable}
\pgfplotsset{compat=newest}
\graphicspath{{graphics/}}
\usepackage{textcomp}

{ \theorembodyfont{\normalfont} 

}

\newcounter{enumctr}

\DeclareFontFamily{U}{mathx}{\hyphenchar\font45}
\DeclareFontShape{U}{mathx}{m}{n}{<-> mathx10}{}
\DeclareSymbolFont{mathx}{U}{mathx}{m}{n}
\DeclareMathAccent{\widebar}{0}{mathx}{"73}

\newcolumntype{C}{>{\centering\arraybackslash}X} 
\DeclareAcronym{alstm}{
    short = ALSTM,
    long  = Attention-based Long Short-Term Memory,
}
\DeclareAcronym{ann}{
    short = ANN,
    long  = Artificial Neural Networks,
}
\DeclareAcronym{aqi}{
    short = AQI,
    long  = Air Quality Index,
}
\DeclareAcronym{auc}{
    short = AUC,
    long  = Area Under the Curve,
}
\DeclareAcronym{cnn}{
    short = CNN,
    long  = Convolutional Neural Networks,
}
\DeclareAcronym{co}{
    short = CO,
    long  = Carbon Monoxide,
}
\DeclareAcronym{co2}{
    short = CO\textsubscript{2},
    long  = Carbon Dioxide,
}
\DeclareAcronym{convlstm}{
    short = Conv-LSTM,
    long  = Convolutional Long Short-Term Memory,
}
\DeclareAcronym{ddpm}{
    short = DDPM,
    long  = Denoising Diffusion Probabilistic Models,
}
\DeclareAcronym{dl}{
    short = DL,
    long  = Deep Learnining,
}
\DeclareAcronym{dpd}{
    short = DPD,
    long  = Dynamic Parcel Distribution,
}
\DeclareAcronym{epa}{
    short = EPA,
    long  = Environmental Protection Agency,
}
\DeclareAcronym{gcn}{
    short = GCN,
    long  = Graph Convolutional Neural Networks,
}
\DeclareAcronym{gdpr}{
    short = GDPR,
    long  = General Data Protection Regulation,
}
\DeclareAcronym{gnn}{
    short = GNN,
    long  = Graph Neural Networks,
}
\DeclareAcronym{gps}{
    short = GPS,
    long  = Global Positioning System,
}
\DeclareAcronym{gru}{
    short = GRU,
    long  = Gated Recurrent Unit,
}
\DeclareAcronym{knn}{
    short = KNN,
    long  = K-Nearest Neighbour,
}
\DeclareAcronym{ldm}{
    short = LDM,
    long  = Latent Diffusion Models,
}
\DeclareAcronym{lgbm}{
    short = LGBM,
    long  = Light Gradient-Boosting Machine,
}
\DeclareAcronym{lstm}{
    short = LSTM,
    long  = Long Short-Term Memory,
}
\DeclareAcronym{mae}{
    short = MAE,
    long  = Mean Absolute Error,
}
\DeclareAcronym{ml}{
    short = ML,
    long  = Machine Learning,
}
\DeclareAcronym{mlp}{
    short = MLP,
    long  = Multilayer Perceptron,
}
\DeclareAcronym{mlr}{
    short = MLR,
    long  = Multiple Linear Regression, 
}
\DeclareAcronym{no}{
    short = NO,
    long  = Nitric Oxide,
}
\DeclareAcronym{no2}{
    short = NO\textsubscript{2},
    long  = Nitrogen Dioxide,
}
\DeclareAcronym{nox}{
    short = NO\textsubscript{x},
    long  = Nitrogen Oxides,
}
\DeclareAcronym{o3}{
    short = O\textsubscript{3},
    long  = Ozone,
}
\DeclareAcronym{pm}{
    short = PM,
    long  = Particulate Matter,
}
\DeclareAcronym{pm1}{
    short = PM\textsubscript{1},
    long  = Particulate Matter 1,
}
\DeclareAcronym{pm10}{
    short = PM\textsubscript{10},
    long  = Particulate Matter 10,
}
\DeclareAcronym{pm2.5}{
    short = PM\textsubscript{2.5},
    long  = Particulate Matter 2.5,
}
\DeclareAcronym{rf}{
    short = RF,
    long  = Random Forest,
}
\DeclareAcronym{rnn}{
    short = RNN,
    long  = Recurrent Neural Networks,
}
\DeclareAcronym{roc}{
    short = ROC,
    long  = Receiver Operating Characteristic,
}
\DeclareAcronym{sbc}{
    short = SBC,
    long  = Single-Board Computer,
}
\DeclareAcronym{smote}{
    short = SMOTE,
    long  = Synthetic Minority Over-sampling Technique,
}
\DeclareAcronym{so2}{
    short = SO\textsubscript{2},
    long  = Sulphur Dioxide,
}
\DeclareAcronym{sr}{
    short = SR,
    long  = Spatial Regression,
}
\DeclareAcronym{stgcn}{
    short = STGCN,
    long  = Spatio-Temporal Graph Convolutional Network,
}
\DeclareAcronym{svm}{
    short = SVM,
    long  = Support Vector Machine,
}
\DeclareAcronym{xgb}{
    short = XGB,
    long  = Extreme Gradient Boosting,
}

\begin{document}
    \title{\LARGE \bf Comparative Analysis of Machine Learning-Based Imputation Techniques for Air Quality Datasets with High Missing Data Rates}

    \author{Sen Yan, \textit{Graduate Student Member, IEEE}, David J. O'Connor, Xiaojun Wang,  \\Noel E. O'Connor, \textit{Member, IEEE}, Alan. F. Smeaton, \textit{Fellow, IEEE} and Mingming Liu, \textit{Senior Member, IEEE}
    
    \thanks{S. Yan, N.E. O'Connor, and M. Liu are with the School of Electronic Engineering and Insight Research Ireland Centre for Data Analytics at Dublin City University, Dublin, Ireland. D.J. O'Connor is with the School of Chemical Sciences at Dublin City University, Dublin, Ireland. X. Wang is with the School of Electronic Engineering and the Entwine Centre at Dublin City University, Dublin, Ireland. A.F. Smeaton is with the School of Computing and Insight Research Ireland Centre for Data Analytics at Dublin City University, Dublin, Ireland. \textit{Corresponding author: Mingming Liu. Email: {\tt mingming.liu@dcu.ie}.}}}

    \maketitle
    \thispagestyle{empty}
    \pagestyle{empty}

    \begin{abstract}\label{sec:abstract}
        \input{data/abstract}
    \end{abstract}


    \IEEEpeerreviewmaketitle
    
    \section{Introduction} 
        \label{sec: intro}
        \input{data/1_introduction}
        
    \section{Literature Review} 
        \label{sec: review}
        \input{data/2_review}

    \section{Data \& Dataset}
        \label{sec: data}
        \input{data/3_data}

    \section{Experiments \& Results}
        \label{sec: exp}
        \input{data/4_experiment}

    \section{Conclusions \& Future Work}
        \label{sec: conclude}
        \input{data/5_conclusion}

    \color{black}
    
    \section*{Acknowledgement} 
        \label{sec: acknowledgement}
        \input{data/acknowledgement}

    \bibliographystyle{IEEEtran}
    \bibliography{reference}

\end{document}

%% file: data/abstract.tex
Urban pollution poses serious health risks, particularly in relation to traffic-related air pollution, which remains a major concern in many cities. Vehicle emissions contribute to respiratory and cardiovascular issues, especially for vulnerable and exposed road users like pedestrians and cyclists. Therefore, accurate air quality monitoring with high spatial resolution is vital for good urban environmental management. This study aims to provide insights for processing spatiotemporal datasets with high missing data rates. \textcolor{black}{In this study, the challenge of high missing data rates is a result of the limited data available and the fine granularity required for precise classification of PM\textsubscript{2.5} levels. The data used for analysis and imputation were collected from both mobile sensors and fixed stations by \textcolor{black}{Dynamic Parcel Distribution}, \textcolor{black}{the Environmental Protection Agency}, and Google in Dublin, Ireland}, where the missing data rate was approximately 82.42\%, making accurate \textcolor{black}{Particulate Matter 2.5} level predictions particularly difficult. Various imputation and prediction approaches were evaluated and compared, including ensemble methods, deep learning models, and diffusion models. External features such as traffic flow, weather conditions, and data from the nearest stations were incorporated to enhance model performance. The results indicate that diffusion methods with external features achieved \textcolor{black}{the highest F1 score, reaching 0.9486 (Accuracy: 94.26\%, Precision: 94.42\%, Recall: 94.82\%), with ensemble models achieving the highest accuracy of 94.82\%}, illustrating that good performance can be obtained despite a high missing data rate.

%% file: data/1_introduction.tex

Urban air pollution is one of the most critical environmental challenges, especially in densely populated cities where vehicle emissions, industrial activities, and residential heating serve as the primary sources of traffic-related pollutants such as \ac{pm} which are major contributors to respiratory and cardiovascular diseases \cite{Yan2023}. Specific \ac{pm} pollutants, such as \ac{pm2.5} and \ac{pm10}, exacerbate the situation by penetrating deep into the lungs, leading to serious health issues, including respiratory problems, cardiovascular disease, lung cancer, and adverse birth outcomes \cite{Henning2024}. The global impact of air pollution is immense, with an estimated 7 million excess deaths annually\footnote{\url{https://www.who.int/health-topics/air-pollution}}, including 3,300 in Ireland alone\footnote{\url{https://www.irishexaminer.com/news/arid-41018408.html}}. As urbanisation increases, real-time, high-resolution, and accurate air quality monitoring systems become crucial to protecting public health.

Vulnerable road users, such as pedestrians and cyclists, are more exposed to the open air in urban environments and typically travel shorter distances, around 500 m \cite{Yan2022}, compared to other modes of transportation. To better protect these users, air quality monitoring systems are essential, particularly for pollutants like \ac{pm2.5}, a key indicator of smog and air quality degradation \cite{Liu2022, Renard2022, Renard2023}. However, predicting \ac{pm2.5} levels is challenging due to limited data availability and the fine granularity required for accurate classification. \textcolor{black}{Some existing solutions, such as the Google Air Quality API\footnote{\url{https://developers.google.com/maps/documentation/air-quality/overview}}, incorporate a large volume of data. However, the data sources are not fully transparent or readily accessible, offering 30-day forecasts without a detailed explanation of the methods or data used for prediction}. In this study, we analyse air quality data from Dublin, Ireland, collected by fixed stations and mobile sensors operated by \ac{dpd} delivery vans, the \ac{epa}, and Google. \textcolor{black}{These datasets provide the most relevant and valuable resources for our analysis. However, the merged dataset still presents} a missing data rate of approximately 82.42\%, making it challenging for reliable predictions. \textcolor{black}{To address this, we focus on implementing effective imputation strategies for robust air pollution management}.

Previous studies have employed techniques for air quality prediction including conventional \ac{ml} models \cite{Arnaudo2020, Gryech2020, Sakib2023}, \ac{dl} approaches \cite{Bekkar2021, Janarthanan2021, Mao2021}, and diffusion models \cite{Chen2024}. Conventional methods, such as \ac{rf} and \ac{knn}, have proven effective for large datasets, while \ac{dl} models, particularly those based on recurrent architectures like \ac{lstm} and \ac{gru}, have demonstrated superior capabilities in capturing complex temporal dependencies. More recently, diffusion models, which account for dynamic spatial relationships and uncertainties in predictions, have emerged as a powerful solution to challenges such as high missing data rates \cite{Chen2024, Dong2024}. This study aims to evaluate and compare the performance of various imputation and prediction methods mentioned above specifically for air quality forecasting. To improve model accuracy, we also incorporate external data sources such as traffic flow data, weather conditions, and nearby monitoring station information.

The main contributions of our work are outlined below:

\begin{itemize}
    \item A comprehensive analysis of air quality data from multiple sources was conducted with fine spatial (0.5 km grid cells) and temporal (hourly) resolution, revealing a high missing data rate of 82.42\% for Dublin.
    \item A comparative experiment was conducted to evaluate various imputation and prediction methods, with \ac{rf} as the best performer, achieving an accuracy of 94.70\%.
    \item The impact of external data sources, such as traffic flow and weather conditions, was evaluated and demonstrated.
    \item Diffusion models, including \ac{ddpm} and \ac{ldm}, were explored regarding their ability to handle spatiotemporal dependencies and missing data.
\end{itemize}

The structure of this paper is as follows. In \autoref{sec: review}, we provide a review of the relevant literature on air quality prediction and imputation methods for spatiotemporal data. The dataset analysis and processing are explained in detail in \autoref{sec: data}. \autoref{sec: exp} presents the prediction problem, experiment setup, and relevant results. Finally, we conclude our work in \autoref{sec: conclude} and discuss future plans for improvements.

%% file: data/2_review.tex
In this section, we provide a literature review of recent work focused on the \ac{ml} methodologies employed to predict air quality and existing studies working on effective strategies used to address high rates of missing data.

\subsection{Air Quality Prediction}

Researchers have applied various \ac{ml} methods in air quality prediction, which can be categorised into four types based on structures of the data and models: classical \ac{ml}-based, station-based, graph-based, and grid-based studies.

Classical \ac{ml}-based studies \cite{Arnaudo2020, Gryech2020, Sakib2023} typically use models like \ac{rf}, \ac{svm}, and regression models such as LR and \ac{mlr}. These models are suitable for structured data with relatively simple feature types and require less computational power, making them effective with simpler datasets. Station-based studies \cite{Bekkar2021, Janarthanan2021, Mao2021, Logothetis2024, Castelli2020, Zhang2021} rely on data from fixed monitoring stations and frequently use \ac{lstm} and its variants (e.g., Conv-LSTM) to capture temporal patterns. However, their effectiveness reduces when applied to broader areas requiring finer spatial resolution. Graph-based studies \cite{Huang2021, Ge2020} map monitoring stations as nodes in a graph, using \ac{rnn} or \ac{gnn} and their variants (e.g., SpAttRNN, MST-GCN) to model complex spatial dependencies. These methods can effectively represent real-world spatial interactions but are computationally expensive as graph complexity grows. Grid-based studies \cite{Chae2021, Abirami2021, Song2022, Le2020, Alex2024} divide the area into spatial grid systems and often combine \ac{cnn} and \ac{lstm} to capture both spatial and temporal features. These studies demand large datasets and computational resources and are sensitive to grid size choices.

Among the existing literature, we found that apart from air pollutants (e.g. \ac{pm2.5}, \ac{pm10}, \ac{no2}, \ac{co}, \ac{so2}, \ac{o3}), weather data such as temperature, relative humidity, and atmospheric pressure are commonly used as external features in air quality prediction. Some studies include traffic data and public vitality metrics. This is consistent with the perspective in \cite{Gualtieri2020, VeeraManikandan2023}, which primarily discussed the impact of wind direction and traffic flow on air pollution. The rates of missing data in the datasets introduced in these studies vary significantly. While some researchers were fortunate to work with datasets without any missing data, others had to manage datasets with missing data rates exceeding 50\%. Given the critical influence of missing data on prediction results, we review existing studies focused on missing data imputation methods below. 

\subsection{Imputation Methods for Spatiotemporal Data}


Missing data imputation and prediction in spatiotemporal datasets have been conducted using approaches including conventional \ac{ml}, \ac{dl}, and diffusion models. We introduce and summarise these methods below.


Conventional \ac{ml} methods, including \ac{knn} \cite{Ahn2022, Jiang2021}, \ac{rf} \cite{Sekuli2020, He2023}, and graph-based approaches \cite{Jiang2021}, are effective for structured data and simpler tasks. For instance, \ac{rf} has been applied successfully in spatiotemporal prediction and imputation tasks, such as precipitation and temperature prediction \cite{Sekuli2020}, and aerosol loading data imputation \cite{He2023}, which demonstrates its ability to handle missing data. \ac{dl} methods, such as \ac{lstm} \cite{Saad2020} and \ac{gru} \cite{Wang2021}, are better suited to capture spatial dependencies in complex data. These models have been used in air pollution prediction, offering improved accuracy compared to statistics-based imputation methods \cite{Drewil2022}. Diffusion models, e.g., \ac{ddpm} \cite{Chen2024} and \ac{ldm} \cite{Dong2024}, represent the latest advancements in data generation within the computer vision field. Researchers have also begun exploring their application in spatiotemporal datasets. For instance, \cite{Chen2024} introduced a novel \ac{ddpm}-based method for air quality prediction, and \cite{Dong2024} proposed an \ac{ldm} to quantify short-term uncertainties in wind power scenario generation. These models can handle uncertainties and dynamic spatial relationships, making them effective for complex real-world data.

In summary, conventional \ac{ml} methods work effectively for simpler tasks, \ac{dl} models offer more advanced capabilities to capture temporal-spatial dependencies, and diffusion models provide advanced capabilities for handling uncertainty. Given these relative strengths, we conducted a comparative study of these methods to evaluate their performance on our data set which is characterised by a high amount of missing data.

%% file: data/3_data.tex
\subsection{Dataset Description}

The data used in this study was collected from three different sources: Google Project Air View Data - Dublin City \cite{Google2023}, \ac{epa} Ireland Archive of Air Quality \cite{EPA2024}, and the \ac{dpd} Air Quality Monitoring Programme\footnote{\url{https://www.rte.ie/news/business/2021/0920/1247720-dpd-launches-air-quality-monitoring-initiative/}}, referred to in this paper as ``Google data'', ``\ac{epa} data'', and ``\ac{dpd} data'', respectively. Google and \ac{epa} data are now open-source and available for download, while \ac{dpd} data is currently shared with universities and will eventually be made publicly available.

\textbf{\textit{Google data}}, sourced from mobile sensors mounted on vehicles, was collected by Google and Dublin City Council from May 2021 to August 2022. It aggregates 1-second interval data into approximate 50-meter road segments, capturing concentrations of \ac{co2} and air pollutants, including \ac{no}, \ac{no2}, \ac{co}, \ac{o3}, and \ac{pm2.5}. \textcolor{black}{This project collected measurements with high spatial resolution but limited frequency, leading to low temporal resolution.}

\textbf{\textit{\ac{epa} data}}, as part of the National Ambient Air Quality Monitoring Programme in Ireland, has been collected from fixed monitoring stations by the Environmental Protection Agency of Ireland since 2017. This dataset contains hourly average concentrations of \ac{so2}, \ac{no2}, \ac{co}, \ac{o3}, and \ac{pm2.5}. \textcolor{black}{It provides comprehensive temporal coverage, but due to the high reliance on fixed stations, it is limited to a few locations.}

\textbf{\textit{\ac{dpd} data}}, collected since September 2021 by \ac{dpd} Ireland and Dublin City Council, was sourced from mobile sensors mounted on 102 \ac{dpd} delivery vehicles and from 22 fixed stations. The sensors are managed by \textcolor{black}{the Pollutrack network \cite{Renard2023_1, Renard2021}, which is funded by \ac{dpd}}, and capture real-time particulate matter concentrations at breathing levels, including \ac{pm1}, \ac{pm10}, and \ac{pm2.5}. \textcolor{black}{The inclusion of mobile and fixed sensors empowered this dataset to provide good spatial and temporal coverage.}

Based on the above, it can be seen that \ac{pm2.5} is the only common air pollutant across all datasets, so in this study, we focused on the only overlapping period from 1 May to 31 July 2022, a total of 92 days, to investigate the factors influencing \ac{pm2.5} levels in urban areas, specifically within inner Dublin. The ``inner Dublin'' region is defined by the bounding coordinates: from (53.3608\textdegree N, -6.3084\textdegree W) in the northwest to (53.3295\textdegree N, -6.2403\textdegree W) in the southeast, including the area near Dublin city centre and the River Liffey, which is one of the busiest areas and the most important transportation hub in the city.

\subsection{Time Series Decomposition}

\begin{figure*}[ht]
    \vspace{-0.1in}
    \centering
    \begin{subfigure}[b]{0.28\linewidth}
        \centering
        \includegraphics[width=\textwidth]{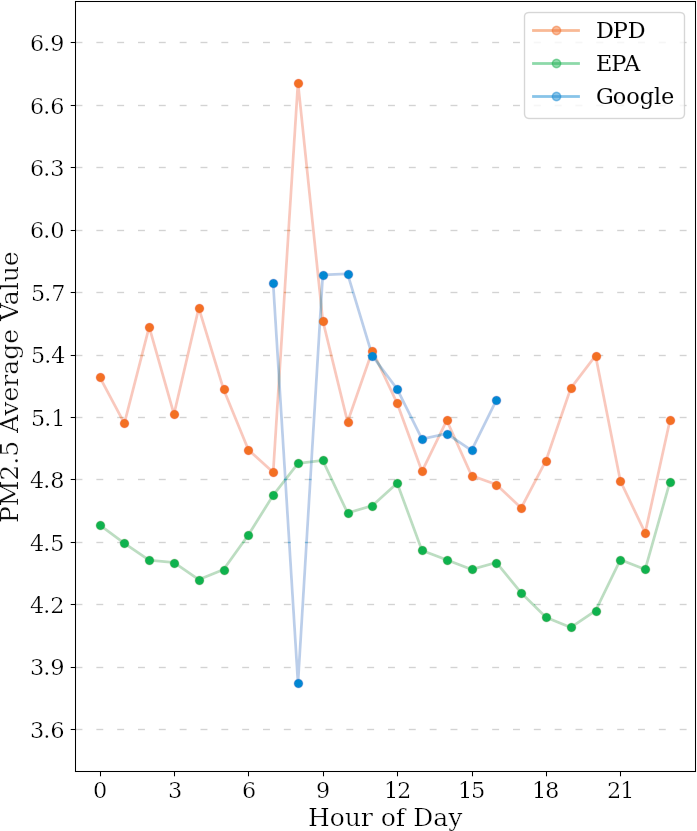}
        \caption{Average \ac{pm2.5} values in 3 datasets.}
        \label{subfig: hourly}
    \end{subfigure}
    \hfill
    \begin{subfigure}[b]{0.68\linewidth}
        \centering
        \includegraphics[width=\textwidth]{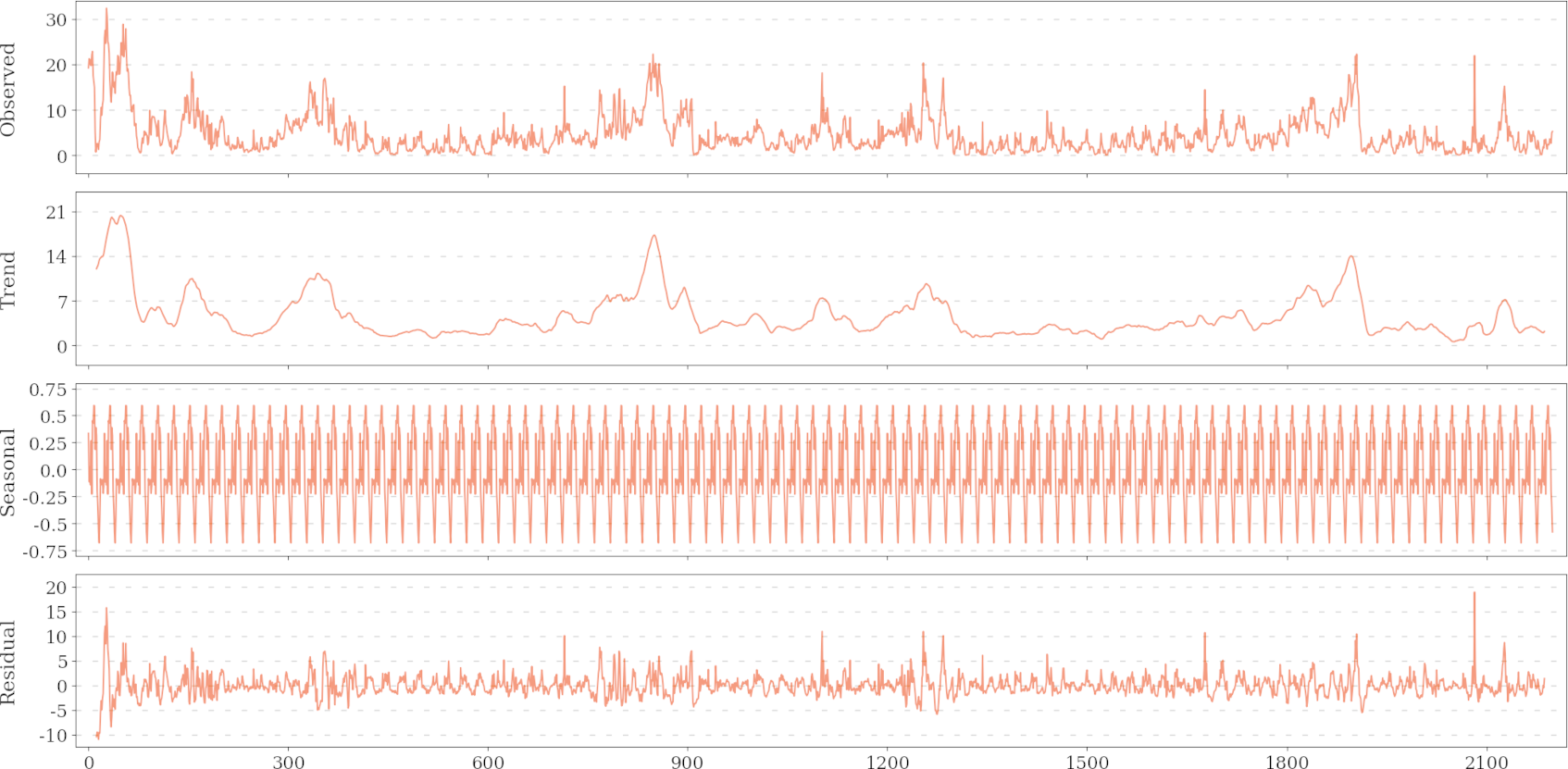}
        \caption{Time series decomposition results of hourly average \ac{pm2.5} values in \ac{dpd} data}
        \label{subfig: decomp}
    \end{subfigure}
    \caption{Comparison of hourly average \ac{pm2.5} values in 3 datasets, with time series decomposition of \ac{dpd} data as an example.}
    \vspace{-0.2in}
    \label{fig: decomposition}
\end{figure*}

We applied time series decomposition to examine the long-term trends, periodic fluctuations, and anomalies of each dataset. The analysis below uses \ac{dpd} data as an example, where the data was grouped and averaged by the hour. The results are presented in \autoref{fig: decomposition}, in which \autoref{subfig: hourly} compares the hourly average \ac{pm2.5} values across the three datasets, revealing significant diurnal variations in \ac{pm2.5} concentrations. Meanwhile, \autoref{subfig: decomp} demonstrates the time series decomposition of the \ac{dpd} data, providing valuable insights into the temporal behaviour of \ac{pm2.5} levels, particularly highlighting a distinct 24-hour seasonal cycle. 

In \autoref{subfig: decomp}. the ``Seasonal'' component highlights consistent daily variations, which may be driven by regular traffic and industrial emissions. This aligns with the hourly average \ac{pm2.5} values shown in \autoref{subfig: hourly}, which demonstrates similar patterns of peaks and troughs. The ``Trend'' component provides a broader view, indicating the impact of long-term influences like weather conditions. Meanwhile, the ``Residual'' component is relatively smooth with occasional spikes, which shows that most of the fluctuations can be explained by the trend and seasonal patterns, but some abnormal events or irregular changes still have some influence on the \ac{pm2.5} levels.

The consistent daily variations mentioned earlier are visualised in detail in \autoref{subfig: hourly}, which shows a sharp increase during the morning rush hours (08:00 to 10:00) and another noticeable rise in the late afternoon (17:00 to 20:00), which align to the evening rush hour. Between these two peaks (11:00 to 16:00) and during the night (21:00 to 07:00 next day), \ac{pm2.5} concentrations stabilise at lower levels, which reflects reduced human activity and traffic. Moreover, the \ac{epa} data exhibits relatively stable \ac{pm2.5} levels throughout the day with only minor fluctuations, possibly because it is collected by fixed monitoring stations. On the other hand, the Google data follows a pattern similar to \ac{dpd}, though with less pronounced peaks and a gap in data from 17:00 to 06:00 the next day, because it is collected by mobile sensors mounted on vehicles. These differences present how each dataset captures distinct characteristics of air quality variations, which could be affected by data collection methods and sensor locations.

Similarly, we examined the hourly average \ac{pm2.5} values by day of the week and the decomposition results in the \ac{dpd} data. The time series decomposition reveals significant periodic fluctuations, indicating that \ac{pm2.5} levels present regular changes over time. While the residual component remains relatively smooth, occasional spikes still represent the occurrence of abnormal events or sudden pollution sources. Notably, \ac{dpd} data shows pronounced peaks on Tuesday and Sunday, with Tuesday reaching a particularly high level, probably due to increased traffic. The Google data demonstrates more rapid changes, peaking sharply on Tuesday before dropping quickly on Wednesday, and lacks data on weekends, possibly because the vehicle-mounted sensors used for collection do not operate during that time. The \ac{epa} data, in contrast, presents a more stable trend with fewer fluctuations.

\subsection{Spatial Visualisation \& Analysis}

\begin{figure}
    \centering
    \includegraphics[width=\linewidth]{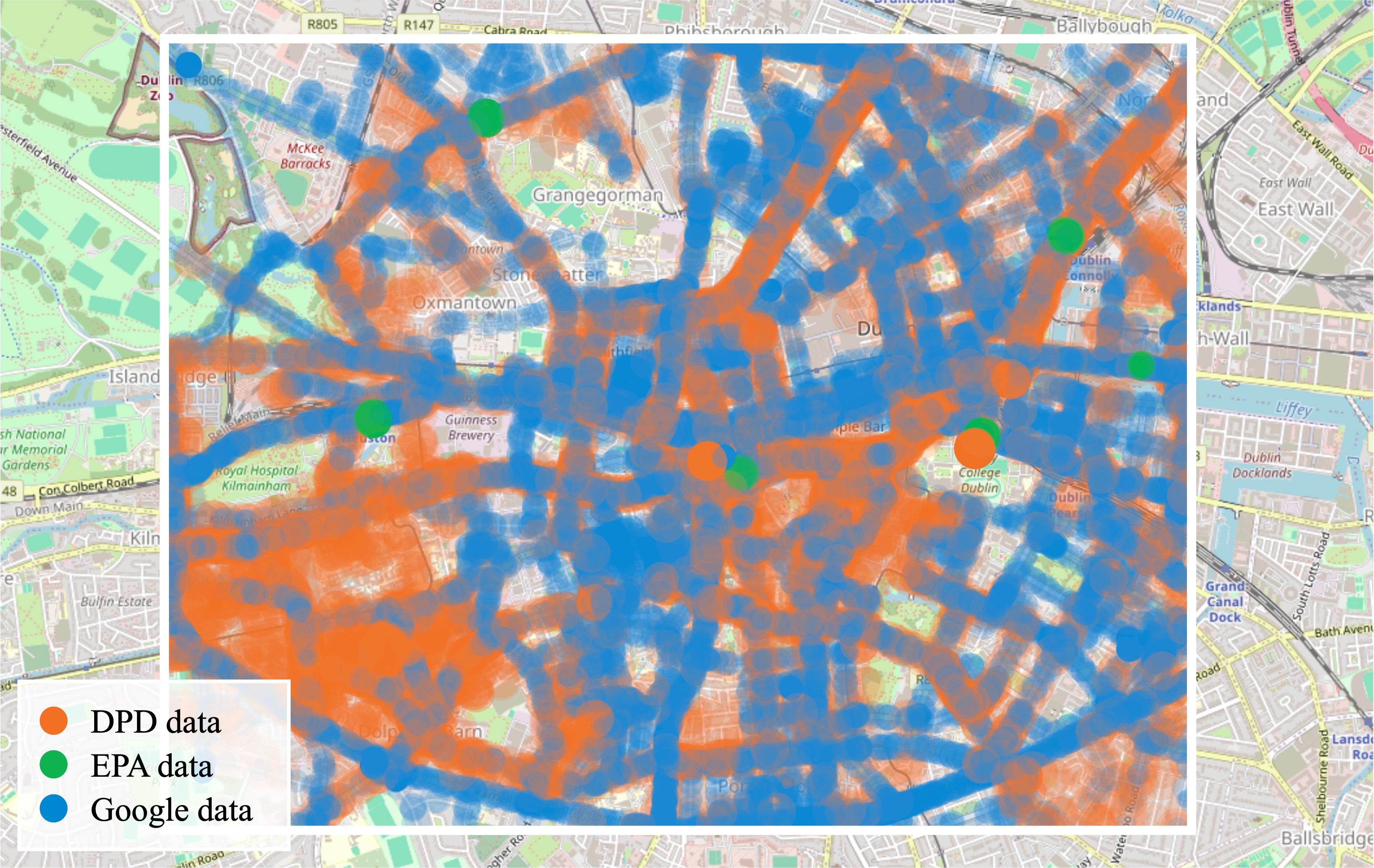}
    \caption{Spatial distribution of \ac{pm2.5} data in three datasets.}
    \vspace{-0.2in}
    \label{fig: spatial}
\end{figure}

To analyse the spatial distribution and pollution intensity within Dublin,  a scatter plot shown in \autoref{fig: spatial} visualises and compares the spatial distribution of air pollution data from three different sources, where the \ac{dpd}, EPA, and Google data are represented by points in orange, green, and blue, respectively. The size of each point corresponds to pollution intensity, with larger points indicating higher pollution levels. Apart from the geographical distribution, it can also be seen from \autoref{fig: spatial} that the EPA data points are sparse and localised, reflecting the fixed locations of those monitoring stations. In contrast, the \ac{dpd} and Google datasets cover a wider spatial area with significant overlap, indicating more extensive data collection across the city. The different point sizes in the \ac{dpd} and Google datasets indicate fluctuations in pollution intensity throughout our focus area and provide a more comprehensive view of air quality dynamics.

\subsection{Data Processing}

In this study, we divided the inner Dublin area into a grid of 500 m $\times$ 500 m cells to enhance spatial resolution. Data within each grid cell was averaged on an hourly basis. This division split inner Dublin into 7 rows and 10 columns, creating 70 grid cells, with an ideal expectation of 2,208 data points per grid (92 days $\times$ 24 hours). However, due to different observation frequencies and relatively low spatial sampling rates in the \ac{dpd}, \ac{epa}, and Google datasets, the missing rates were quite high, at 89.61\%, 91.51\%, and 98.68\%, respectively. \textcolor{black}{Subsequently, these three datasets were merged by averaging the available values within each cell, resulting in a reduced missing rate of} 82.32\%, which shows the benefit of combining datasets to improve data completeness. However, based on a dataset with such a high missing rate, any downstream tasks will be challenging, so in this study, we address this issue by using \ac{ml}-based imputation methods to fill missing values.

We initially extracted the day of the week and the hour of the day from the timestamp as additional temporal features. For spatial information, we applied one-hot encoding to the latitudes and longitudes to make the model more focused on the grid system. Moreover, we extracted the observations from the nearest two fixed stations within our datasets and calculated their distances to each target cell centre, since they are suggested to be useful in air pollution prediction \cite{Sekuli2020}.

In addition, we included external features such as traffic information and weather conditions, which have also been shown to significantly influence air pollution levels \cite{Gualtieri2020, VeeraManikandan2023}. The traffic data, obtained from an open dataset \cite{SCATS2022_1, SCATS2022_2}, contains two key metrics: \textit{Sum Volume}, representing the total traffic volume in the preceding hour, and \textit{Avg Volume}, indicating the average traffic volume per 5-minute interval in the preceding hour, while weather conditions, collected from the Irish Meteorological Service \cite{MET2024}, include various climate indicators, such as \textit{Precipitation} and \textit{Air Temperature}. Both traffic and weather data were averaged on an hourly basis.

Lastly, we divided the \ac{pm2.5} values into several categories based on the European \ac{aqi} Levels Scale from \cite{OpenWeather2024}. According to this scale, the hourly \ac{pm2.5} concentrations are divided into five groups: Very Low (0-15 $\text{\textmu g/m}^3$), Low (15-30 $\text{\textmu g/m}^3$), Medium (30-55 $\text{\textmu g/m}^3$), High (55-110 $\text{\textmu g/m}^3$), and Very High (over 110 $\text{\textmu g/m}^3$), where the first group aligns with the World Health Organisation's global air quality guideline level for \ac{pm2.5} in \cite{WHO2021}. However, after examining our dataset, we found that only one observation exceeded 110 $\text{\textmu g/m}^3$, leading to a significant class imbalance. \textcolor{black}{Furthermore, the data points are heavily imbalanced across the five classes, with ratios of 25,901:1,227:167:17:1, which would significantly impact model performance. To address this}, we initially combined the last two categories, reducing the task to a four-class classification problem, \textcolor{black}{and then employed the \ac{smote} algorithm \cite{Chawla2002} on our training set to reduce the impact of class imbalance on model training by synthesising samples from the minority classes}. The processed dataset is presented in \autoref{tab: dataset}. The \textit{label} column represents our prediction target, where missing values are filled with ``$-1$'', and the rest of the entries are labelled with their corresponding class numbers.


\begin{table*}
    \caption{Sample Data of \ac{pm2.5} Records from the Processed Dataset.}
    \vspace{-0.1in}
    \label{tab: dataset}
    \begin{tabularx}{\linewidth}{@{\extracolsep{\fill}}c c c c c c c c c c c c c c c c c c}
        \toprule
        \textbf{wkd} & \textbf{h} & \textbf{lat} & \textbf{lon} & \textbf{t\_sum} & \textbf{t\_avg} & \textbf{w\_p} & \textbf{w\_t} & \textbf{w\_w} & \textbf{w\_d} & \textbf{w\_v} & \textbf{w\_r} & \textbf{w\_m} & \textbf{s\_dis\_1} & \textbf{s\_val\_1} & \textbf{s\_dis\_2} & \textbf{s\_val\_2} & \textbf{label} \\
        \midrule
        2 & 14 & 1 & 0 & 0 & 0 & 0.0 & 16.1 & 12.3 & 8.8 & 11.3 & 61 & 1020.2 & 2.8380 & 2.8196 & 1.9002 & 3.4328 & 0 \\
        3 & 11 & 1 & 0 & 0 & 0 & 0.0 & 16.5 & 13.2 & 10.5 & 12.7 & 67 & 1024.2 & 2.8380 & 2.3190 & 1.9002 & 1.7567 & 0 \\
        0 & 10 & 1 & 1 & 0 & 0 & 0.0 & 14.4 & 11.7 & 9.0 & 11.5 & 70 & 1015.0 & 2.8380 & 5.8679 & 1.9002 & 5.7697 & 0 \\
        2 & 10 & 1 & 1 & 0 & 0 & 0.0 & 13.7 & 12.2 & 10.8 & 12.9 & 82 & 1007.6 & 2.8380 & 0.7776 & 1.9002 & 4.0162 & 2 \\
        3 & 10 & 1 & 1 & 0 & 0 & 0.1 & 16.8 & 15.2 & 14.0 & 16.0 & 83 & 1016.7 & 2.8380 & 0.3800 & 1.9002 & 0.1703 & 0 \\
        \bottomrule
    \end{tabularx}
    \begin{tablenotes}\footnotesize
        \item[*] \textbf{wkd} -- Weekday, \textbf{h} -- Hour, \textbf{lat} -- One-hot Coded Latitude, \textbf{lon} -- One-hot Coded Longitude, \textbf{t\_sum} -- Total Traffic Volume, \textbf{t\_avg} -- Average Traffic Volume, \textbf{w\_p} -- Precipitation Amount (mm), \textbf{w\_t} -- Air Temperature ($^\circ\text{C}$),  \textbf{w\_w} -- Wet Bulb Temperature ($^\circ\text{C}$),  \textbf{w\_d} -- Dew Point Temperature ($^\circ\text{C}$), \textbf{w\_v} -- Vapour Pressure (hPa), \textbf{w\_r} -- Relative Hmidity (\%), \textbf{w\_m} -- Mean Sea Level Pressure (hPa), \textbf{s\_dis\_1} -- Station 1 Distance, \textbf{s\_val\_1} -- Station 1 Observation, \textbf{s\_dis\_2} -- Station 2 Distance, \textbf{s\_val\_2} -- Station 2 Observation, \textbf{label} -- \ac{pm2.5} level.
    \end{tablenotes}
    \vspace{-0.2in}
\end{table*}

%% file: data/4_experiment.tex
This section introduces the experiment setups and provides a comprehensive analysis of the experiment results.



\subsection{Experiment Setup}

In this section, we applied various \ac{ml} models and compared their performance in a missing data imputation task. We categorised the models into three groups: conventional \ac{ml} models, \ac{dl} models, and diffusion models. The detailed setups for each model category are introduced below.
        
\subsubsection{Conventional \ac{ml} Models} We applied \ac{knn}, \ac{rf}, \ac{xgb}, and \ac{mlp} models to our dataset without two different feature setups: one used only the initial features, i.e., \textit{wkd}, \textit{h}, \textit{lat}, \textit{lon}, and \textit{label} columns as shown in \autoref{tab: dataset}, and the other included more external features such as \textit{t\_sum}, \textit{w\_t}, and \textit{s\_dis\_1}. This allowed us to evaluate model performance both with and without the inclusion of external features. The dataset was split into a training set and a test set with a ratio of $4:1$, and the \ac{smote} algorithm was applied to the training set to address class imbalance. For the \ac{knn} model, we used 4 nearest neighbours to determine the predictions. In the \ac{rf} model, 10 estimators were employed to run 10 jobs in parallel. The \ac{xgb} model used the classifier from the \ac{xgb} library with a gradient-boosted tree booster, running 300 jobs in parallel. For the \ac{mlp} model, we built two hidden layers with 82 and 83 neurons respectively, using the hyperbolic tangent activation function. The model was optimised with the Adam optimiser with an adaptive learning rate.

\subsubsection{\ac{dl} Models} An \ac{lstm}-based model is designed as a classifier in our study, i.e., \ac{lstm} (c) in \autoref{tab: results}. It consists of two \ac{lstm} layers with 128 and 64 units, respectively, using L2 regularisation to prevent overfitting. Batch normalisation and dropout layers are applied between \ac{lstm} and dense layers to stabilise training and improve generalisation, with dropout rates set to 0.30. Two fully connected dense layers with 64 and 32 units are applied before the softmax output layer, which predicts class probabilities. The model is compiled with the Adam optimiser, with a learning rate of $1 \times e^{-4}$, and evaluated using accuracy, precision, recall, and F1 score metrics. 

In addition to serving as a classifier, we also developed an \ac{lstm} model for time series prediction. Its architecture is similar to the \ac{gru} model, with each model consisting of two recurrent layers, followed by batch normalisation, dropout layers with a dropout rate of 0.6, and a final dense layer for classification. The \ac{lstm} model uses 64 and 16 units in its two \ac{lstm} layers, while the \ac{gru} model utilises 64 and 32 units in its \ac{gru} layers. Both models apply ReLU activation in the dense layers and use softmax for multi-class classification. Both models were optimised using the Adam optimiser with an initial learning rate of $1 \times e^{-3}$. Evaluation metrics are the same as those used for conventional \ac{ml} models.

\subsubsection{Diffusion Models} In this study, both the \ac{ddpm} and \ac{ldm} models were designed for time series classification using convolutional architectures to extract spatial features. \ac{ddpm} in our work was designed with a U-Net structure to process time serial data. The forward diffusion process simulates noise-contaminated data by gradually adding noise to the data at multiple time steps. For each time step $t$, the process of adding noise can be described as:

\begin{equation}
    q(x_t | x_{t-1}) = \mathcal{N}(x_t; \sqrt{1 - \beta_t} x_{t-1}, \beta_t I)
\end{equation}

\noindent where $q(x_t | x_{t-1})$ is the conditional probability of $x_t$ given $x_{t-1}$, and $x_t$ is the noise-added data at time step $t$. $\mathcal{N}(x_t; \mu, \sigma)$ is the distribution of data $x_t$ under a normal distribution with mean $\mu$ and covariance $\sigma$. $\beta_t$ here means the intensity of the noise added at time step $t$, and $I$ denotes the identity matrix. Similarly, the reverse diffusion process can be described as:

\begin{equation}
    p_\theta(x_{t-1} | x_t) = \mathcal{N}(x_{t-1}; \mu_\theta(x_t, t), \Sigma_\theta)
\end{equation}

\noindent where $p_\theta(x_{t-1} | x_t)$ is the learned reverse process, approximating the true reverse distribution. $\mu_\theta(x_t, t)$ is the mean of U-Net predictions, and $\Sigma_\theta$ is the covariance matrix. 

\ac{ldm} transfers the diffusion process from the original data space $x_t$ to the latent space $z_t$ via an encoder $f_{encoder}(.)$, which is a convolutional network in our study, and recovers the original data from the latent space the diffusion process via a decoder $f_{decoder}(.)$ through the reverse convolution operation.

These two models both include convolutional layers with L2 regularisation and dropout, followed by a softmax output layer for multi-class classification, and each model was trained using the Adam optimiser with a learning rate of $1 \times e^{-3}$ and was evaluated based on accuracy, precision, recall, and F1 score. Additionally, both models utilise 100 timesteps to balance effective noise modelling and computational efficiency. The primary distinction between the two lies in their diffusion process. \ac{ddpm} adds noise directly to the input data and employs a U-Net architecture with skip connections, while \ac{ldm} applies the diffusion process in a latent space, compressing the input before decoding it back into the original space.

\subsection{High Accuracy of Ensemble Models}

The experimental results are summarised in \autoref{tab: results}. It can be seen that ensemble learning models, such as \ac{rf} and \ac{xgb}, outperform both traditional machine learning and deep learning models across various evaluation metrics. \ac{rf} achieved the highest accuracy of 94.82\% and an F1 score of 0.9440, indicating a strong balance between precision and recall. \ac{xgb} closely followed, with accuracy scores of 93.89\%. These results highlight the effectiveness of ensemble methods in handling complex datasets with missing values and spatiotemporal patterns, especially when external features are added, which helped them better capture the factors influencing \ac{pm2.5} levels.

Moreover, the considerable precision-to-recall ratios across ensemble models demonstrate their ability to classify positive instances while maintaining precision well. For instance, \ac{rf} achieved a precision of 0.9442 and a recall of 0.9482, minimising false positives and capturing most true positives.

\begin{table}[htbp]
    \caption{Prediction Results of \ac{pm2.5} Level (Classification).}
    \label{tab: results}
    \vspace{-0.1in}
    \begin{tabularx}{\linewidth}{@{\extracolsep{\fill}}c c c c c}
        \toprule
        \textbf{Method} & \textbf{Accuracy} & \textbf{Precision} & \textbf{Recall} & \textbf{F1 score} \\
        \midrule
        LDM       & 0.9426 & \textbf{0.9589} & 0.9426 & \textbf{0.9486} \\
        RF (wf)   & \textbf{0.9482} & 0.9442 & \textbf{0.9482} & 0.9440 \\
        XGB (nf)  & 0.9389 & 0.9273 & 0.9389 & 0.9331 \\
        XGB (wf)  & 0.9204 & 0.9191 & 0.9204 & 0.9197 \\
        RF (nf)   & 0.9268 & 0.9107 & 0.9268 & 0.9187 \\
        DDPM      & 0.9176 & 0.9181 & 0.9162 & 0.9171 \\
        KNN (wf)  & 0.8863 & 0.9263 & 0.8863 & 0.9059 \\
        MLP (wf)  & 0.9038 & 0.9067 & 0.9038 & 0.9052 \\
        KNN (nf)  & 0.8131 & 0.8993 & 0.8131 & 0.8540 \\
        LSTM (c)  & 0.8444 & 0.8659 & 0.7994 & 0.8313 \\
        LSTM (s)  & 0.8067 & 0.8104 & 0.8012 & 0.8058 \\
        MLP (nf)  & 0.7206 & 0.9102 & 0.7206 & 0.8044 \\
        GRU       & 0.7563 & 0.7691 & 0.7052 & 0.7358 \\
        \bottomrule
    \end{tabularx}
    \begin{tablenotes}\footnotesize
        \item (c): classifier \phantom{bbbbbbbbbbbbbbbbbbb} (nf): with no external features
        \item (s): serial prediction \phantom{bbbbbbbbbbbbb} (wf): with external features
    \end{tablenotes}
    \vspace{-0.2in}
\end{table}

\subsection{High F1 Score of Diffusion Models}

\begin{figure*}[ht]
    \vspace{-0.1in}
    \centering
    \begin{subfigure}[b]{0.49\textwidth}
        \centering
        \includegraphics[width=\textwidth]{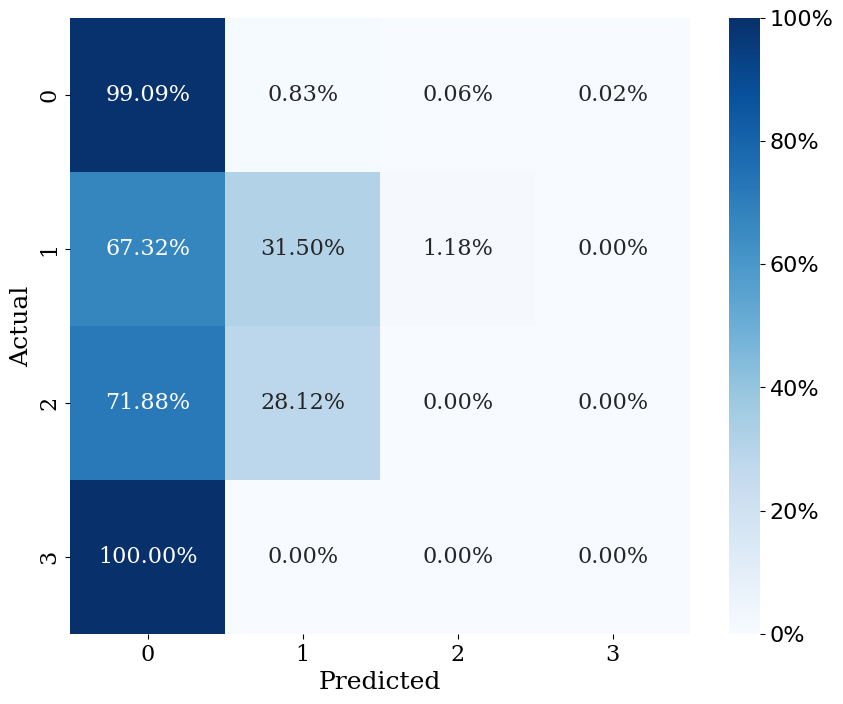}
        \caption{Confusion matrix for \ac{rf}.}
        \label{subfig: rf_cm}
    \end{subfigure}
    \hfill
    \begin{subfigure}[b]{0.49\textwidth}
        \centering
        \includegraphics[width=\textwidth]{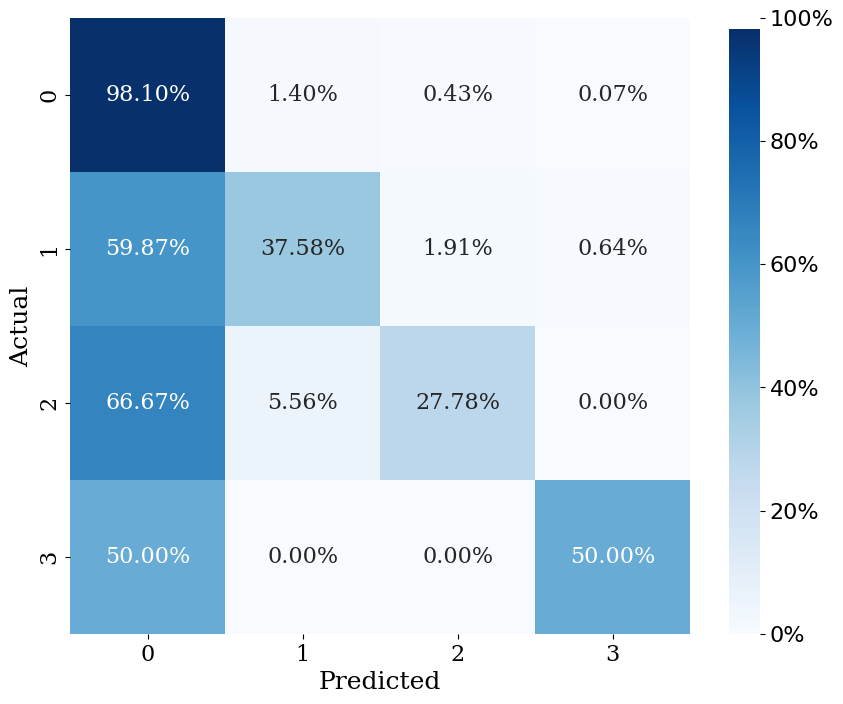}
        \caption{Confusion matrix for \ac{ldm}.}
        \label{subfig: ldm_cm}
    \end{subfigure}
    \hfill
    \begin{subfigure}[b]{0.49\textwidth}
        \centering
        \includegraphics[width=\textwidth]{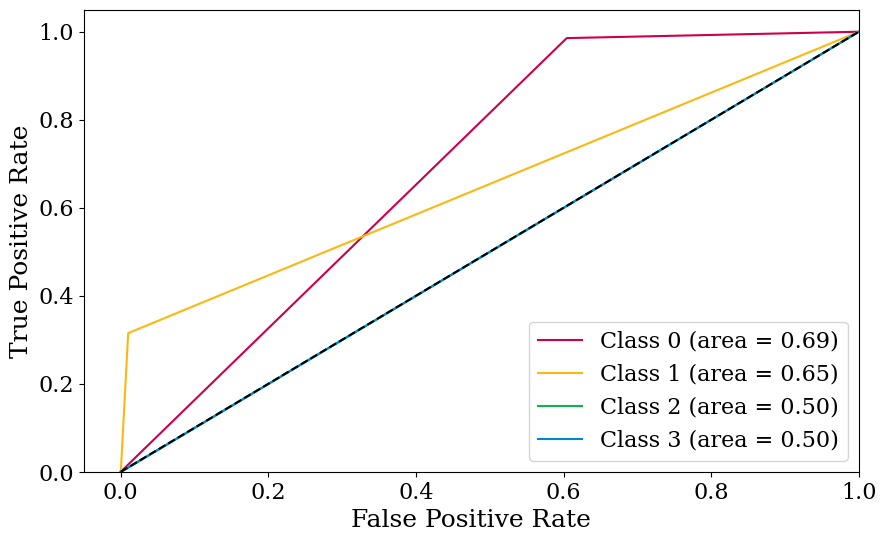}
        \caption{\ac{roc} curves for \ac{rf}.}
        \label{subfig: rf_roc}
    \end{subfigure}
    \hfill
    \begin{subfigure}[b]{0.49\textwidth}
        \centering
        \includegraphics[width=\textwidth]{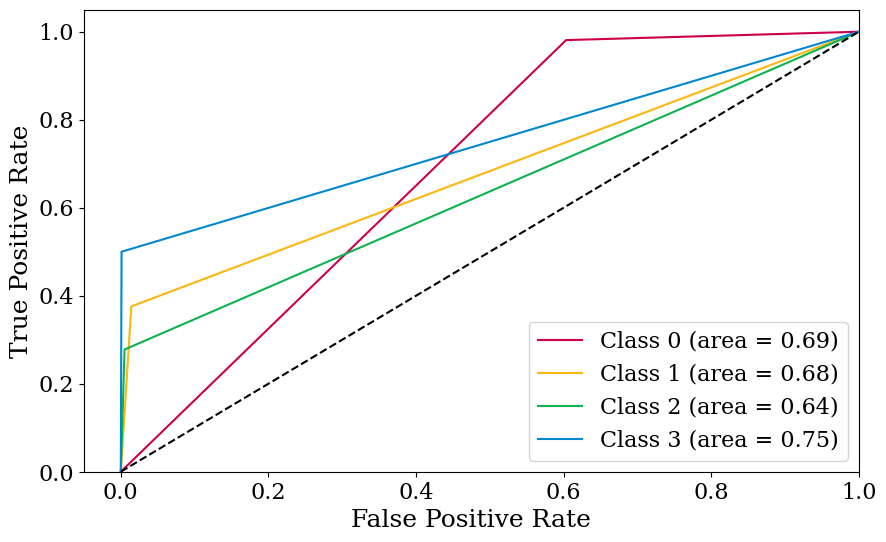}
        \caption{\ac{roc} curves for \ac{ldm}.}
        \label{subfig: ldm_roc}
    \end{subfigure}
    \caption{Confusion matrix and \ac{roc} curves for \ac{rf} and \ac{ldm}.}
    \label{fig: cm_and_roc}
    \vspace{-0.2in}
\end{figure*}

Diffusion models such as \ac{ddpm} and \ac{ldm} demonstrated strong performance in the PM2.5 classification task. \ac{ldm} achieved an accuracy of 94.26\% with an F1 score of 0.9486, and \ac{ddpm} followed with an accuracy of 91.76\% and an F1 score of 0.9171. To further explore and compare the performances of \ac{ldm} against the best performer in accuracy, we analysed and visualised their confusion matrices and \ac{roc} curves in \autoref{fig: cm_and_roc}. 

It is clear that \ac{rf} achieved the highest accuracy of 99.09\% for classifying low \ac{pm2.5} levels (class 0 in \autoref{subfig: rf_cm}), but faced difficulties in distinguishing higher pollution levels, particularly between classes 1 and 2. The \ac{roc} in \autoref{subfig: rf_roc} curves showed that \ac{rf} performed best for class 0, with an \ac{auc} of 0.69, while the \ac{auc}s for classes 1 and 2 were lower, at 0.65 and 0.50, respectively. In contrast, \ac{ldm} demonstrated more balanced results across all \ac{pm2.5} levels. While the accuracy of \ac{ldm} for class 0 was slightly lower at 95.22\%, it performed significantly better for classes 1 and 2, with accuracies of 79.62\% and 72.22\%, respectively. The \ac{roc} curves further confirmed its superior ability to discriminate between pollution levels, particularly for class 2, where it achieved an \ac{auc} of 0.86.

\subsection{Impact of External Features}

The inclusion of external features had a significant impact on model performance. For instance, the external features increased the accuracy of \ac{knn} from 81.31\% to 88.63\%. Similarly, the accuracy of \ac{mlp} improved significantly from 72.06\% to 90.38\% with the addition of external data. These results highlighted the influence of broader environmental and human activity factors in determining \ac{pm2.5} levels. However, the inclusion of external features slightly decreased the accuracy of \ac{xgb} from 93.89\% to 92.04\%. This suggests that \ac{xgb} might be powerful enough to handle feature interactions effectively without needing additional external inputs.

The results indicate that \ac{rf} provides the best accuracy for predicting \ac{pm2.5} levels, while diffusion models achieved the best F1 score, which indicates their strong potential in handling complex spatiotemporal data. These models effectively incorporate external features, allowing them to account for the environmental and human activity impacts on \ac{pm2.5}. Additionally, balancing precision and recall is essential in this classification task, especially for identifying higher \ac{pm2.5} levels, which indicate poorer air quality. Some models, such as \ac{knn} and \ac{mlp}, exhibited high precision rates but struggled with recall rates, resulting in lower F1 scores. In contrast, ensemble learning models and diffusion models achieved a much better balance between precision and recall, enabling them to accurately capture both positive and negative cases across the four categories.

%% file: data/5_conclusion.tex
In this study, we conducted a comprehensive analysis of air quality data collected in Dublin, Ireland showing a significant missing data rate of 82.42\% for a merged dataset, which results in challenges for accurately predicting \ac{pm2.5} levels. To address this, we evaluated and compared various imputation and prediction methods, including conventional models, \ac{dl} approaches, and diffusion models. Experimental results indicate that \ac{ldm} achieved the highest F1 score (0.9486) and \ac{rf} achieved the highest accuracy (94.82\%) when external features such as traffic flow and weather conditions were also included. This demonstrates their potential to handle complex spatiotemporal dependencies and high missing data rates. These results illustrate the importance of considering external factors in air quality prediction and offer insights into addressing spatiotemporal data challenges under conditions of significant missing data.

\textcolor{black}{Future research could explore several directions. Further refining diffusion models would enhance their performance, and extending the analysis to other pollutants and geographic regions would help validate the generalisability of the findings. Inspired by \cite{Yan2023}, another avenue could involve integrating this work with route planning and providing real-time route recommendations based on dynamic air pollution information for vulnerable road users.}

%% file: data/acknowledgement.tex
The authors would like to thank Sidra Aleem from the SFI Centre for Research Training in Machine Learning (ML-Labs) at Dublin City University and Eric Arazo Sánchez from CeADAR, Ireland’s Centre for Applied Artificial Intelligence at University College Dublin, for their support and suggestions on diffusion models. We also appreciate Pranjnay Bhardwaj and Shantanu Singh from Dublin City University for their assistance in the literature review. This publication has emanated from research conducted with the financial support of Taighde Éireann — Research Ireland under Grant number \textit{21/FFP-P/10266} and \textit{SFI/12/RC/2289\_P2}. The \ac{dpd} dataset is provided by the Air Quality Monitoring Programme operated by \ac{dpd} Ireland (URL: \url{https://www.dpd.ie/sustainability}).